\documentclass[runningheads]{llncs}
\usepackage{graphicx}

\usepackage{tikz}
\usepackage{comment}
\usepackage{amsmath,amssymb} 
\usepackage{color}

\begin{document}
\pagestyle{headings}
\mainmatter
\def\ECCVSubNumber{}  

\title{RPR-Net: A Point Cloud-based Rotation-aware Large Scale Place Recognition Network} 

\titlerunning{Abbreviated paper title}

\author{
Zhaoxin Fan \inst{1} \and
Zhenbo Song\inst{3} \and
Wenping Zhang\inst{1} \and
Hongyan Liu\inst{2} \and
Jun He\inst{1} \thanks{111} \and
Xiaoyong Du\inst{1}}
\authorrunning{Fan et al.}
%
\institute{Key Laboratory of Data Engineering and Knowledge Engineering of MOE,
School of Information, Renmin University of China, 100872, Beijing, China \and
Department of Management Science and Engineering, Tsinghua University, 100084, Beijing, China \and
School of Computer Science and Engineering, Nanjing University of Science and Technology, 210094, Nanjing, China\\
\email{fanzhaoxin@ruc.edu.cn, songzb@njust.edu.cn, wpzhang@ruc.edu.cn, hyliu@tsinghua.edu.cn,  hejun@ruc.edu.cn, duyong@ruc.edu.cn}\\
}

\maketitle

\begin{abstract}
Point cloud-based large scale place recognition is an important but challenging task for many applications such as Simultaneous Localization and Mapping (SLAM). Taking the task as a point cloud retrieval problem, previous methods have made delightful achievements. However, how to deal with catastrophic collapse caused by rotation problems is still under-explored. In this paper, to tackle the issue, we propose a  novel Point Cloud-based \textbf{R}otation-aware Large Scale \textbf{P}lace \textbf{R}ecognition \textbf{Net}work (RPR-Net). In particular, to solve the problem, we propose to learn rotation-invariant features in three steps. First, we design three kinds of novel Rotation-Invariant Features (RIFs), which are low-level features that can hold the rotation-invariant property. Second, using these RIFs, we design an attentive module to learn rotation-invariant kernels. Third, we apply these kernels to previous point cloud features to generate new features, which is the well-known SO(3) mapping process. By doing so, high-level scene-specific rotation-invariant features can be learned. We call the above process an Attentive Rotation-Invariant Convolution (ARIConv). To achieve the place recognition goal, we build RPR-Net, which takes ARIConv as a basic unit to construct a dense network architecture. Then, powerful global descriptors used for retrieval-based place recognition can be sufficiently extracted from RPR-Net. Experimental results on prevalent datasets show that our method achieves comparable results to existing state-of-the-art place recognition models and significantly outperforms other rotation-invariant baseline models when solving rotation problems.

\end{abstract}

\section{Introduction}
Autonomous Driving (AD) and Simultaneous Localization and Mapping (SLAM) are becoming increasingly more important in some practical applications. Moreover, large scale place recognition often acts as the spine of them. It plays an indispensable role in a SLAM system or an autonomous driving system. Specifically, on the one hand, place recognition could provide a self-driving car with accurate localization information in a high definition map (HD Map). On the other hand, the recognition result is always used for loop-closure detection \cite{campos2021orb,memon2020loop} in a SLAM system.

\begin{figure}[h]
	\centering  
	\includegraphics[width=0.99\textwidth]{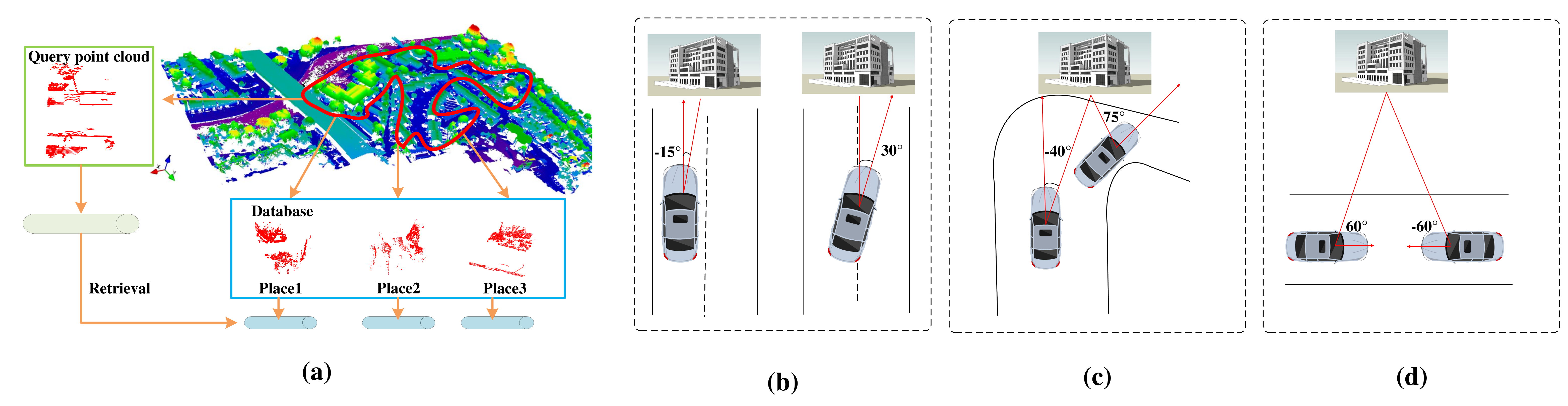}  
	\vspace{-0.15in}
	\caption{Task definition and motivation. (a) an illustration of point cloud-based large scale place recognition. (b) change lanes causes rotation problems. (c) turning a corner causes rotation problems. (d) driving from different direction causes rotation problems. } 
	\label{task}
    \vspace{-0.25in}
\end{figure} 
Image and point cloud are the two kinds of  most frequently used data formats for large scale place recognition.  However, the former is proven to be sensitive to illumination changes, weather changes, etc., making methods \cite{arandjelovic2016netvlad,yu2019spatial} based on it difficult to achieve robust performance. In contrast, point cloud  scanned by LiDAR is more reliable because  it is calculated from reflected lasers, which is less sensitive to the above environmental changes. Therefore, for performance purpose, we study point cloud-based large scale place recognition in this paper.

Fig.\ref{task} (a) illustrates a pipeline of point cloud-based large scale place recognition methods: 1) A HD Map of an area is firstly pre-built as a strong prior. 2) Then, the HD Map is uniformly divided into a database of submaps with accurate localization information. 3) When a car or a robot travels the area, the scanned point cloud at hand should be compared with each of these submaps to find the most similar location-available submap to  determine the location of the car. The first two steps are easy to do offline, while the third step, to find the most similar sample from a large amount of topologically variable point clouds in an on-the-fly retrieval manner, is harder to achieve. Apparently, the place recognition problem is essentially the foundation of such a challenging retrieval problem.

For the point cloud retrieval problem, challenges lay in how to encode point clouds into discriminative and robust global descriptors for similarity calculation. In recent years, many deep learning-based point cloud encoding models \cite{uy2018pointnetvlad,zhang2019pcan,liu2019lpd,sun2020dagc,fan2020srnet,komorowski2021minkloc3d,xia2021soe,fan2021svt} have been proposed and have achieved acceptable recognition performances. However, we observe that nearly all existing methods are facing a serious common defect: they are suffered from catastrophic collapse caused by rotation problems, i.e., when  the scanned point clouds are rotated compared to the pre-defined submaps, their performance would significantly drop (evidenced in Fig. \ref{compare} and Tabel \ref{tablebaseline}).  Undoubtedly, this would  make the algorithm unreliable and  cause ambiguities for decision making. What's more, safety risks would rise  if these models are deployed to self-driving cars and robots.  What makes things worse is that rotation problems are very common in practical scenarios. As shown in Fig. \ref{task} (b) (c) and(d), when a vehicle or robot changes lanes, turns a corner or drives from different directions, the scanned point cloud is equivalent to being rotated relative to the submaps in the HD map. Hence, to increase reliability and safety, it is very necessary and important to take the \emph{rotation problems} into consideration when designing point cloud-based place recognition models.

In this paper, we propose to solve the problem by learning rotation-invariant global descriptors. In particular, we propose a novel point cloud-based rotation-aware large scale place recognition model named RPR-Net. To power RPR-Net with the rotation-invariant ability, we first design a kind of Attentive Rotation-Invariant Convolution named ARIConv, which learns rotation-invariant features in three steps:

 First, taking a point cloud as input,  three kinds of Rotation-Invariant Features (RIFs) named Spherical Signals (SS), Individual-level Local Rotation-Invariant Features (ILRIF) and Group-level Local Rotation-Invariant features (GLRIF) are extracted.  The three kinds of features hold the rotation-invariant  property of the point cloud from three different perspectives: invariance of individual points in spherical domain,  invariance of individual-level relative position in a local region, and invariance of point distribution in a local region. All the three kinds of features are low-level features which are representative. Second, taking RIFs as input, we propose an attentive module to learn rotation-invariant convolutional kernels. In the attentive module, momentous RIFs that are more relevant to a specific scene are adaptively highlighted, such that the most significant rotation-invariant related hidden modes can be enhanced. To do so, we implement a RIF-wise attention in the high-level latent space. Through this way, the model could learn how to re-weight different RIFs' mode to generate more representative convolutional kernels to achieve the scene-specific SO(3) mapping process. Third, we apply the learned convolutional kernels to features of previous layers to obtain high-level features of the current layer. Due to that the above process is a SO(3) mapping process that can effectively inherit the rotation-invariant property of both RIFs and  the input features, both the learned kernels and the corresponding convolved point cloud features would perpetually be strictly rotation-invariant.

To complete the large scale place recognition task, we further propose a dense network architecture to build RPR-Net using ARIConvs, which benefited from learning rotation-invariant features as well as strengthening semantic characteristic of scenes.  After trained, RPR-Net would predicted robust and discriminative global rotation-insensitive descriptors from raw point clouds. Then, recognition-by-retrieving can be achieved.

We have conducted extensive experiments following the evaluation setting of top rotation-invariant related papers \cite{you2020pointwise,you2021prin}. Experimental results  show that our proposed model can achieve comparable results to existing state-of-the-art large scale place recognition models. Meanwhile, our model also significantly outperforms other strong rotation-invariant baselines. What's more, the advantage of our novel proposed model is even more remarkable when dealing with rotation problems of point clouds. Specifically, when the point clouds are rotated at different rotation levels, recognition accuracy of our model is almost constant, yet all state-of-the-art place recognition competitors fail to work.

Our contributions can be summarized as: 
\begin{itemize}
\item We propose a novel model named RPR-Net, which, to the best of our knowledge, is the first strictly rotation-invariance dense network designed for point cloud-based large scale place recognition.

\item We propose an Attentive Rotation-Invariant Convolution operation, which learns rotation-invariant by mapping low-level rotation-invariant features into attended convolutional kernels.

\item  We achieve the state-of-the-art performance when dealing with rotation problems and achieve comparable results with most of existing methods on several original non-rotated benchmarks.

\end{itemize}

\section{Related Works}
\subsection{Rotation-invariant convolution}
In our work, the attentive rotation-invariant convolution is a key component. In early years, deep learning models like PoinetNet \cite{qi2017pointnet}, PointNet++ \cite{qi2017pointnet2} and DGCNN \cite{wang2019dynamic} try to  use a T-Net \cite{qi2017pointnet} module to make models less sensitive to rotation. It learns a rotation matrix to transform  a raw point cloud into canonical coordinates. Though operating on raw point clouds, T-Net is not strictly rotation-invariant. In fact, it is challenging to achieve strictly rotation-invariant using a normal convolution operation \cite{rao2019spherical,chen2019clusternet,dym2020universality}.  Methods like Spherical cnns \cite{cohen2018spherical} and \cite{esteves2018learning} propose to project 3D meshes onto their enclosing spheres for learning global rotation-invariant features. They are not suitable for raw point cloud-based methods and in-efficient. To solve the issue, Sun et al. \cite{sun2019srinet} propose SRINet to learn point projection features that are rotation-invariant. Zhang et al. \cite{zhang2019rotation} design RIConv, which utilizes low-level rotation-invariant geometric features like distances and angles to learn  convolutional weights for learning rotation-invariant features. Kim et al. \cite{kim2020rotation} propose to utilize graph neural networks to learn rotation-invariant representations. Li et al. \cite{li2020rotation}  introduce a region relation convolution to encode both local and non-local informations. Their main goal is alleviating the inevitable global information loss caused by  rotation-invariant representations. Recently, You et al. \cite{you2020pointwise,you2021prin} propose PRIN, where an adaptive sampling strategy and a 3D spherical voxel convolution operation are introduced to learn point-wise rotation-invariant features. These works, though are strictly rotation-invariant, are tailored for point cloud classification or segmentation.  They either cannot learn sufficient scene-level features, or are too memory expensive, making them  not applicable to the large scale place recognition task. In our work, the ARIConv is designed to construct the dense RPR-Net, which can learn both powerful geometric features and powerful semantic features as well as keeping the rotation-invariant property. To our knowledge, we are the first to design  strictly rotation-invariant convolution for large scale place recognition.

\begin{figure*}
	\centering  
	\includegraphics[width=0.99\textwidth]{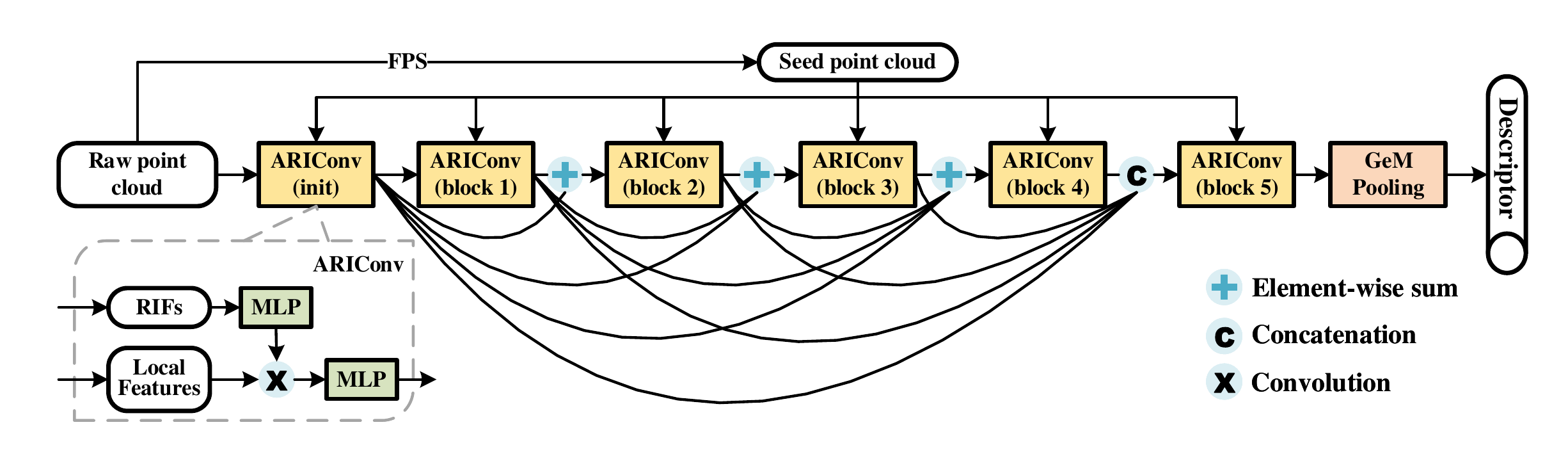}  
\vspace{-0.2in}
	\caption{Illustration of network architecture. } 
	\label{network}
	
\end{figure*}
\subsection{Large scale place recognition}
According to available data formats, large scale place recognition methods can be categorized into image-based methods and point cloud-based methods. Both methods share the same scheme: first encode scenes into global descriptors and then use K-Nearest-Neighbors (KNNs) algorithm to retrieve similar location-available scenes. Commonly, designing better global descriptors attracts  more research attention. As described before, image-based methods\cite{arandjelovic2016netvlad,yu2019spatial} are sensitive  to changes of illumination, weather, season, etc. Thus, they are not realiable.  Compared with them, point cloud-based methods are more reliable. PointNetVLAD \cite{uy2018pointnetvlad} is the first point cloud-based  deep learning method that achieves successful place recognition results. It uses PointNet \cite{qi2017pointnet} and NetVLAD \cite{arandjelovic2016netvlad} to learn global descriptors. PointNetVLAD only takes advantage of global features and neglects the importance of extracting local/non-local features, therefore, a line of works propose different ways to improve it. Methods like \cite{liu2019lpd,sun2020dagc,fan2020srnet} employ Graph Convolutional Networks \cite{wang2019dynamic} to capture better local features, while methods like \cite{zhang2019pcan,sun2020dagc,fan2020srnet,xia2021soe} use self-attention mechanism \cite{woo2018cbam} to  capture better non-local features to improve performance. There are also some methods \cite{komorowski2021minkloc3d,fan2021svt} propose to  voxelize unstructured point clouds into regular voxels for learning global descriptors, which also achieve acceptable place recognition results. Evidenced by our experimental results, these existing methods are greatly suffered from rotation problems. In this paper, we propose RPR-Net and ARIConv, which demonstrate notable advantages over existing  point cloud-based large scale place recognition methods.

\section{Methodology}
\subsection{Overview}
\textbf{Problem statement:} Our goal is to achieve place recognition when a car or robot travels around an area as shown in Fig. \ref{task} (a). Given a HD Map $M$ of an area, we first divide it into a database of submaps represented as point clouds, denoted by $D=\{s_1, s_2, \cdots,  s_m\}$, where $m$ is the number of submaps. Each submap is attached with its unique localization information. When the car/robot travels around the area, it would scan a  query point cloud $q$. We need to find where the car/robot is in the HD Map utilizing $D$ and $q$. To achieve the goal, we formulate it as a retrieval problem, defined as: $s^*=\phi (\varphi(D|\Theta),\varphi(q|\Theta))$, where $\varphi$ is the deep learning model encoding all point clouds into global descriptors, $\Theta$ is model parameters, $\phi$ is a KNN searching algorithm, and $s^*$ is the most similar submap to $q$. After $s^*$ is retrieved, we regard $q$  and $s^*$ share the same localization information, i.e., the place recognition task at current location is achieved. In this paper, we focus on designing a novel deep learning model $\varphi$ to learn robust and discriminative rotation-invariant point cloud descriptors. We formulate $\varphi$ as our RPR-Net.

\noindent \textbf{Network architecture:}
We first introduce the network architecture of  RPR-Net. To learn representative descriptors, we expect RPR-Net be equipped with the following powers: 1) It should be rotation-invariant. 2) Descriptors learned from it can represent the scene geometry well. 3) High-level scene semantics can be effectively extracted.  With all three aspects being satisfied, the final learned features would be powerful enough for scene description and simlarity calculation. To achieve the three goals, we propose ARI-Conv (detailed later) and a dense network architecture as shown in Fig \ref{network}. Specifically, 6 ARIConvs and one GeM Pooling \cite{radenovic2018fine} layer  are used to constitute the skeleton of the network. The former is used to learn high-level point-wise rotation-invariant features, while the later is used to aggregate the these features into a global descriptor. Note that ARI-Conv mainly explores how to learn rotation-invariant utlizing  geometric characteristics. Therefore, the skeleton is principally equipped the two first two kinds powers, while learning high-quality scene semantics is still hard to achieve if ARIConv layers are simply stacked.  To this end, to learn the third power, we adopt the  \emph{densely connected}  idea \cite{huang2017densely,liu2019densepoint,yu2020deep} to  design a dense network architecture, whose benefits has been demonstrated in many works. To implement, for ARIConv block 1 to 4, we choose to take the element-wise sum of all previous layers' features as their input.  Then, we concatenate outputs of the 4 layers and regard it as the input of the last ARIConv layer (ARIConv block 5), which learns the final point-wise rotation-invariant features. We use element-wise sum in block 1 to 4 rather than use concatenation like in \cite{fan2020srnet} mainly due to that  element-wise sum is much more efficient. Through such a process, powerful and discriminative rotation-invariant  global descriptors can be obtained and later used for efficient retrieval-based place recognition. Next, we introduce the ARIConv and its attributes in detail.

\subsection{Low Level Rotation Invariant Features}
ARIConv's duty is to learn point-wise rotation-invariant features. As mentioned before, its process consists of three steps. The first step is to extract low-level Rotation-Invariant Features (RIFs) from a point cloud, which will be used to learn rotation-invariant kernels and features later.  In our work, each point's RIFs are extracted in a local neighbourhood level. Specifically, given a input point cloud $p \in R^{N \times 3}$, we first use Farthest-Point-Sampling (FPS) to sample a seed point cloud $p_s \in R^{N_s \times 3}$. Then, for each point of $p_s$,  we search the $K$ nearest neighbor of it from $p$, which can be  defined as a matrix of dimension $N_s \times K \times 3$. And the RIFs are extracted from each $K \times 3$ local group.  For each points, we extract three kinds of RIFs: Spherical Signals, Individual-level Local Rotation-Invariant Features and Group-level Local Rotation-Invariant Features. We choose these RIFs due to three reasons.: 1) They are extracted in a local region, which can better represent scene details and defend outliers. 2) Most of them are low-level geometry scalers, so that they can preserve the scene geometry better. 3) They are strictly rotation-invariant and are easy to be calculated. These advantages would benefit the final retrieval task. Next, we introduce them in detail.

\begin{figure}[h]
	\centering  
	\vspace{-0.2in}
	\includegraphics[width=0.8\textwidth]{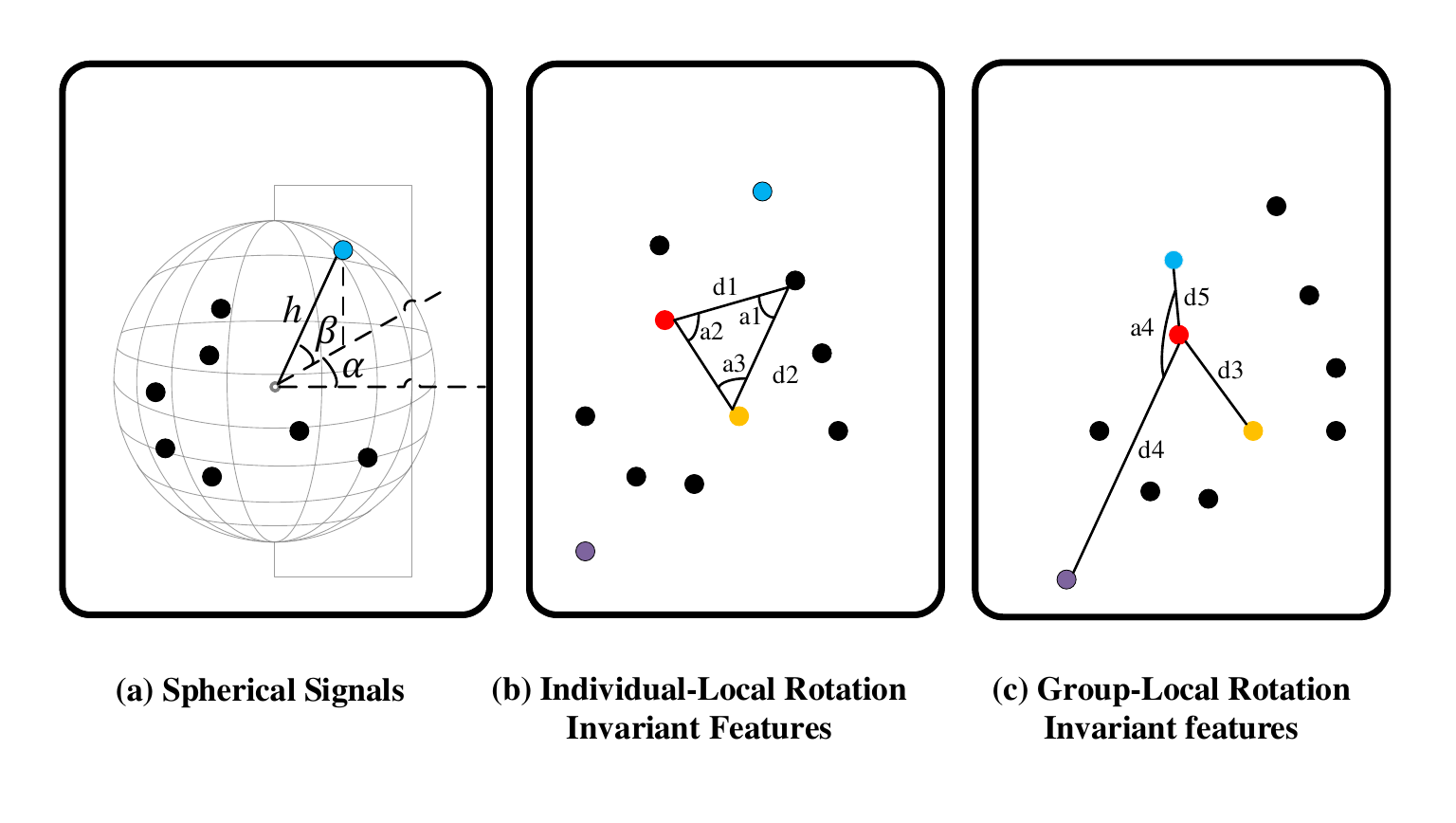} 
	\vspace{-0.2in}
	\caption{Illustration of three kinds of RIFs.} 
	\vspace{-0.2in}
\label{features}
\end{figure}

\noindent  \textbf{Spherical Signals} Suppose $x_i$ is any point in point cloud $p$. The spherical signals are obtained by transforming $x_i$ into its spherical coordinate: $(s(\alpha,\beta),h)$, where $s(\alpha,\beta) \in S^2$ denotes its location in the unit sphere, and $h \in H=[0,1]$ denotes the radial distance to the coordinate origin, as shown in Fig. \ref{features} (a). It has been proven in many works \cite{you2021prin,cohen2018spherical,esteves2018learning} that the spherical signals constructed from the unit spherical space is rotation-invariant. Therefore, we also adopt spherical signals as features for learning rotation-invariant kernels in our work. For convenient, we use the method proposed in SPRIN \cite{you2021prin} to build the spherical signal $f$ operating on the raw sparse point cloud. Besides, considering that the radial distance $h$ is also rotation-invariant since it is a scalar and the coordinate origin will never be changed, we also adopt it as part of the rotation-invariant spherical signals. In total, for an input point cloud, we can get a dimension of $N_s \times K \times 2$  spherical signals.

\noindent
\textbf{Individual-level Local Rotation-Invariant Features} Besides spherical signals, another representation that is constant rotation-invariant is the relative distance between two points as well as the angles between two vectors. Therefore, inspired by RIConv \cite{zhang2019rotation}, we extract five  features for each point as shown in Fig. \ref{features} (b), where the red point is defined as the center point (or seed point) $x_i$ of a $K \times 3$ local group, the yellow point is the mean center of all points in the  local group, and black points are the $K$-nearest neighbors of $x_i$. It is obviously that for each individual black point,  no matter how the point cloud is rotated, features $[d_1, d_2,a_1,a_2,a_3]$ will never be changed, i.e., they are rotation-invariant, where $d_i$ means distance and $a_i$ means angles. In total, for the whole point cloud, we can get a dimension of $N_s \times K \times 5$ individual-level local rotation-invariant features. Note that, for each individual black point (neighbor point) in Fig. \ref{features} (b), the five features are unique and different from that of other points.

\noindent
\textbf{Group-level Local Rotation-Invariant Features}  Individual-level local rotation-invariant features have described the invariance of each individual points in a local group, while the global invariance of the local group itself has not been characterized. For a group, we find that what insensitive to rotation is the  \emph{mean/min/max}   property  between  all points. Therefore, as shown in Fig. \ref{features} (c), we extract $[d_3, d_4,a_5,a_4]$ as our group-level local rotation-invariant features to describe the characteristic of the local group itself. In Fig. \ref{features} (c), the red point is the center point (or seed point) $x_i$ of a $K \times 3$ local group, the yellow point is the mean center of all points in the local group and the blue and purple point are the closest and farthest neighbors of the seed point $x_i$ among the $K$ neighbors, respectively. For $N_s$ local groups, we can  get $N_s  \times 4$ group-level local rotation-invariant features. In order to keep the dimension of it to be consistent with the other two kinds of RIFs, we repeat it $K$ times and finally form a $N_s  \times K \times  4$ feature matrix. From another point of view, it means all points in a local group share the same GLRIFs. Note both the ILRIFs and GLRIFs are strong in expressing scene geometry besides  learning  rotation-invariant features, benefited from the inherent attributes of relative distances and relative angles between different points.

\subsection{Attentive Rotation-Invariant Convolution Operation}
Through the above process, we can extract a $N_s  \times K \times  11$ RIFs matrix. Taking the matrix as input, the second and the third steps are to generate rotation-invariant kernels and apply these kernels to input features, i.e., to conduct the convolution operation.

\noindent
\textbf{Kernels Generation:}  We take three aspects into mind when learning convolutional kernels. First, they should be rotation-invariant. Second, they should be equipped with strong scene representation ability. Third, they should be computational-efficient. To meet these requirements, we choose to use a shared MLP \cite{qi2017pointnet} to map the RIFs into convolutional kernels. Since the MLP processes each point individually, it is obviously rotation-invariant and it has been proven in many works \cite{qi2017pointnet} that MLP is both effective and efficient. Utilizing the shared MLP, we can get a $N_s  \times K \times  C_{in} C_{out}$ kernel matrix $\kappa$. The kernel matrix can be then used  for convolving input features of the seed point cloud $p_s$.  However, simply using kernels learned by a shared MLP neglects the fact that each RIF may play a different role in generating the kernel. This motivates us that we should treat different RIFs with different importance. To this end, we decide to introduce a RIF-wise attention mechanism in the latent space to solve this problem. Specifically, in $\kappa$, we propose to multiply  each element of $C_{in} C_{out}$ channels with a learned attention score to highlight more important RIFs  in an implicit mode. The attention mechanism can be easily implemented by:
\begin{equation}
\kappa= \hat{\kappa} \cdot  sigmoid(fc(avg\_pool(\hat{\kappa})))
\end{equation}

where $\hat{\kappa}$ is the kernels before being attended. Finally, we get a $N_s  \times K \times  C_{in} C_{out}$ attentive kernel matrix $\kappa$. We reshape it into $N_s  \times K \times  C_{in} \times  C_{out}$ and use it to obtain more high-level point-wise rotation-invariant point cloud features later. We call the above process an attentive module. Note again, the purpose of the attentive module is not to attend individual RIF, but to promote the model  to learn how to re-weight different RIFs' latent mode, so that more representative convolutional kernels can be generated.

\noindent
\textbf{Convolution Operation:} After the convolutional kernels are learned, high-level output features are obtained by applying the kernels to the input features using a convolution operation. Suppose the input features of the seed point cloud is $f_s \in R^{N_s \times C_{in}}$. We use the $K$-nearest neighbor indices at the previous step to construct a feature matrix $f_{sg}$.  The dimension  of $f_{sg}$ is $N_s \times K \times C_{in}$. Taking $f_{sg}$ and $\kappa$ as input, the convolution operation for each point can be implemented as: 

\begin{equation}
f_n= \sum_{k=1}^K \sum_{c_{in}=1}^{C_{in}}\kappa(n,k,c_{in})f_{sg}(n,k,c_{in})
\end{equation}

where $\kappa(n,k,c_{in}) \in C_{out}$ is vector, $f_{sg}(n,k,c_{in})$ is scalar  and $f_n$ is the convolved features. Totally, we will get a  $N_s \times C_{out}$ final convolved features matrix. Then, we use a shared MLP to get the final output rotation-invariant features $f_{so}\in R^{N_s \times C_{out}}$.  Since the kernels are rotation-invariant, the above convolutional process is actually a SO(3) mapping process.

\noindent Till now, we have introduced the RPR-Net's dense network achitecture and the proposed ARIConv. As stated before, we use a  GeM Pooling \cite{radenovic2018fine} layer to aggregate  learned point-wise features into a final global descriptor. Since GeM Pooling only operating on each point individually and  its average pooling operation is channel wise, there is no doubt that it is rotation-invariant. Therefore, our whole network architecture is rotation-invariant. One may also notice that each ARIConv layer takes a seed point cloud as its input. For efficiency, we let all ARIConv layers share the same seed point cloud.


\begin{figure*}[h]
	\centering  
	\vspace{-0.15in}
	\includegraphics[width=0.8\textwidth]{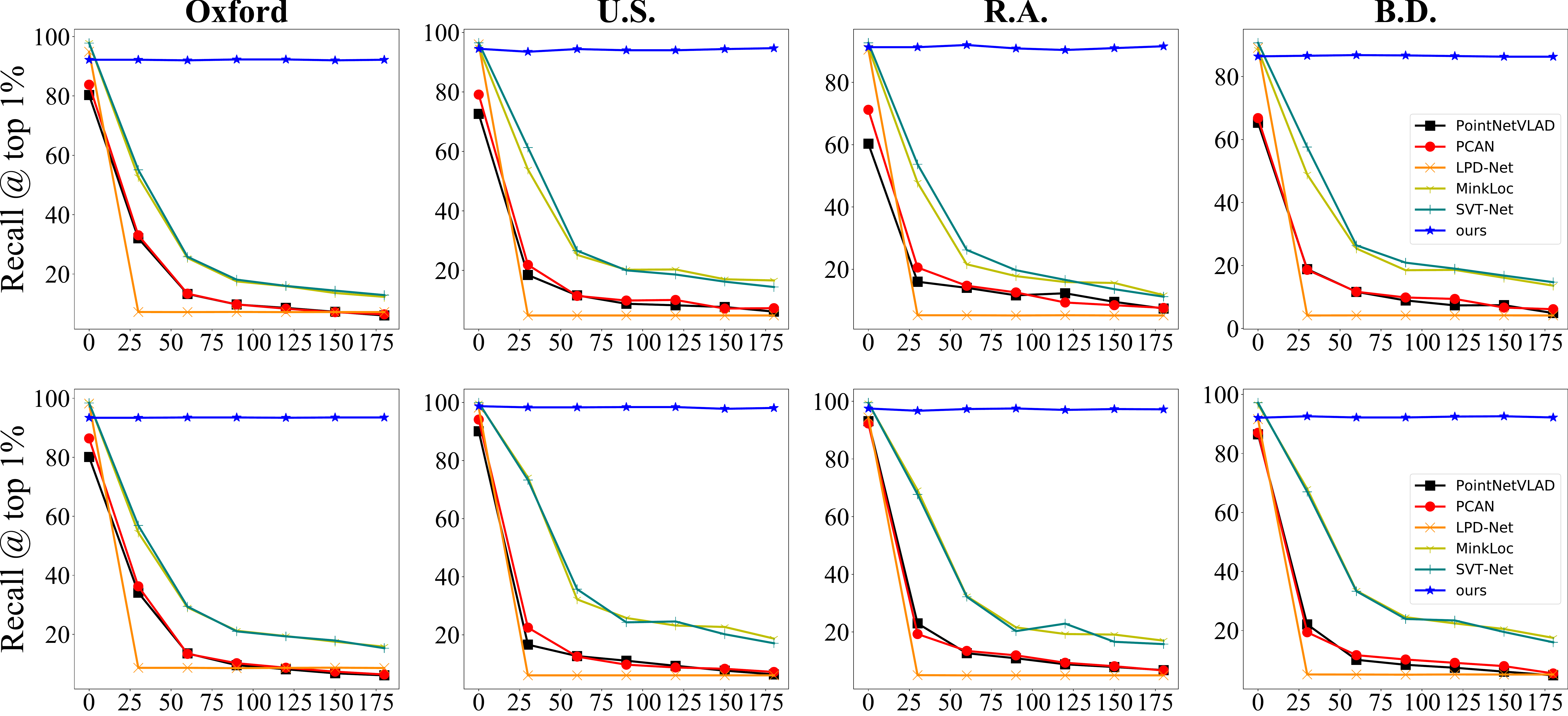} 
	\vspace{-0.1in} 
	\caption{Comparison of our method with state-of-the-art place recognition methods at different rotation levels. The top row is the result at the baseline training setting; the bottom row is the result at the refined training setting. The values along the horizontal axis represents different maximal rotation degrees.} 
	\label{compare}
	\vspace{-0.2in}
\end{figure*}

\section{Experiments}

\subsection{Dataset}
To verify the effectiveness of our method, we use the benchmark datasets proposed by \cite{uy2018pointnetvlad}, which are now the most authoritative and representative datasets for point cloud-based large scale place recognition. The benchmark consists of four datasets: one outdoor dataset called Oxford generated from Oxford RobotCar \cite{maddern20171} and three in-house datasets: university sector (U.S.), residential area (R.A.) and business district (B.D.). Each point cloud in these datasets contains 4096 points after removing the ground plane.   Each point cloud is shifted and rescaled to be zero mean and inside the range of
[-1, 1]. Therefore, the impact of translation is filtered. Each dataset contains 21,711/400/320/200 submaps for training and 3,030/80/75/200 scans for testing for Oxford., U.S., R.A. and B.D., respectively. During training, point clouds are regarded as correct matches if they are at maximum 10m apart and wrong matches if they are at least 50m apart. During testing, the retrieved point cloud is regarded as a correct match if the distance is within 25m between the retrieved point cloud and the query scan.

\begin{table*}[h]
	\centering
	\resizebox{0.9\textwidth}{!}{
	\begin{tabular}{c|c|c|c|c|c|c|c|c}
		\hline
		& \multicolumn{4}{|c|}{{\bf Avg recall at top 1\%(w/o rotation)}} & \multicolumn{ 4}{|c}{{\bf Avg recall at top 1 (w/o rotation)}} \\
		\hline
		& {\bf Oxford} & {\bf U.S.} & {\bf R.A.} & {\bf B.D.} & {\bf Oxford} & {\bf U.S.} & {\bf R.A.} & {\bf B.D.} \\
		
		{\bf PointNetVLAD} &       80.3 &       72.6 &       60.3 &       65.3 &       81.0     &     77.8       &     69.8       &     65.3  \\
		
		{\bf PCAN} &       83.8 &       79.1 &       71.2 &       66.8 &      83.8      &    79.1        &   71.2   &   66.8  \\
		
		{\bf LPD-Net} &       94.9 &         96.0 &       90.5 &       89.1 &       86.3 &         87.0 &       83.1 &       82.3 \\
		
		{\bf Minkloc3D} & {\bf 97.9} &         95.0 &       91.2 &       88.5 &         93.0 &       86.7 &       80.4 &       81.5 \\
		
		{\bf SVT-Net} &       97.8 & {\bf 96.5} & {\bf 92.7} & {\bf 90.7} & {\bf 93.7} & {\bf 90.1} & {\bf 84.3} & {\bf 85.5} \\
		
		{\bf Ours} & 92.2 & 94.5 & 91.3 & 86.4 &   81.0 & 83.2 & 83.3 & 80.4 \\
		\hline
		\hline
		{\bf } & \multicolumn{ 4}{|c|}{{\bf Avg recall at top 1\%(w/ rotation)}} & \multicolumn{ 4}{|c}{{\bf Avg recall at top 1 (w/ rotation)}} \\
		
		\hline
		& {\bf Oxford} & {\bf U.S.} & {\bf R.A.} & {\bf B.D.} & {\bf Oxford} & {\bf U.S.} & {\bf R.A.} & {\bf B.D.} \\
		\hline
		{\bf PointNetVLAD} &     5.0      &     5.8      &     5.8       &    4.0        &     1.6        &     2.1       &      2.6       &      2.5      \\
	
		{\bf PCAN} &      5.2      &       5.7     &     4.5       &    4.4        &      1.7      &   1.7         &     1.6       &     2.3       \\
	
		{\bf LPD-Net} &     7.2       &    4.8        &     5.2       &     4.2       &     2.2       &     1.4       &    1.1        &    2.2        \\
		
		{\bf Minkloc3D} &    11.9        &   14.1     &     11.7   &     13.3   &  4.7   &     6.8       &    6.5 &       8.5     \\
		
		{\bf SVT-Net} &    12.1    &    14.0    &     9.6   &    12.9   &   4.8   &    6.4    &   5.1    &    8.6  \\
		
		{\bf Ours} & {\bf 92.2} & {\bf 93.8} & {\bf 91.4} & {\bf 86.3} & {\bf 81.1} & {\bf 83.3} &   {\bf 82.0} &   {\bf 80.0} \\
		\hline
	\end{tabular} } 
	
	\caption{Quantitative comparison of our methods with sate-of-the-art place recognition models at the baseline training setting. Top rows are results on the original datasets. Bottom rows are results when both submaps and query point clouds are randomly rotated along three axis.  }
	\vspace{-0.35in}
	\label{tablebaseline}
	
\end{table*}
\subsection{Implementation Details}
We use Triplet loss to train our model due to its effectiveness,
following \cite{komorowski2021minkloc3d,fan2021svt}. The output feature dimension of all ARIConv layers is 64 except the final layer \emph{ARIConv block 5}, which is 256.  Therefore, the dimension $l$ of the final global descriptor is also 256. The number of neighbors $K$ in each local group is set as $32$.  Following previous works \cite{komorowski2021minkloc3d,fan2021svt}, we adopt two training  settings. The baseline training setting only uses the training set of the Oxford dataset to train models and the models are trained for 40 epochs. The  refined training setting addtionally adds the training set of U.S. and B.D except using the Oxford dataset and the models are trained for 80 epochs.  We use average recall at top 1\% and average recall at top 1 as evaluation metrics following all previous works. We implement our model using  PyTorch \cite{paszke2017automatic} platform and optimize it using  the Radam optimizer \cite{liu2019variance}. Random jitter, random translation, random points removal and random erasing augmentation are adapted for data augmentation during training.  To  verify our model's  superiority in solving problems caused by rotation, we also follow existing top rotation-invariant related papers  \cite{you2020pointwise,you2021prin} to randomly rotate point clouds for evaluation, which is a fair setting since LiDAR point cloud is collected around. More details can be found in the \textbf{supplementary material}.

\subsection{Results}
In this subsection, to verify the effectiveness and robostness of our model, we first quantitatively compare our method with state-of-the-art large scale place recognition methods including PointNetVLAD \cite{uy2018pointnetvlad}, PCAN \cite{zhang2019pcan},  LPD-Net \cite{liu2019lpd}, MinkLoc3D \cite{komorowski2021minkloc3d} and SVT-Net \cite{fan2020srnet}. We also compare our method with strong rotation-invariant baseline models RIConv \cite{zhang2019rotation} and SPRIN  \cite{you2021prin}. Qualitative verification and complexity analysis are also included.


\noindent
\textbf{Quantitative comparison with place recognition models:} In Fig. \ref{compare}, we compare our method with existing top place recognition methods at both the baseline training setting and the refined training setting. The point clouds are randomly rotated along the gravity axis by different rotation levels. It is the most common case that point clouds are rotated along the gravity axis compared to submaps in practical scenarios. The compared methods are trained by using random rotation as data augmentation.   Note that methods like LPD-Net have tried to used T-Net to solve the rotation problems. It can be seen that all compared methods fail even  the  point clouds are only rotated by a small degree, while performance of our method is almost constant, which significantly demonstrates that our model is more robust towards rotation compared to other methods. The poor performance of other methods verifies that data augmentation and T-Net are not sufficient enough to make the trained models be strong enough to deal with rotation problems. In contrast, our RPR-Net, for the first time, learns strictly rotation-invariant global descriptors, which are much more robust and reliable.  
\begin{table}
\centering
\resizebox{0.9\textwidth}{!}{
\begin{tabular}{c|c|c|c|c|c|c|c|c}
		\hline
		& \multicolumn{4}{|c|}{{\bf Avg recall at top 1\%(w/o rotation)}} & \multicolumn{ 4}{|c}{{\bf Avg recall at top 1 (w/o rotation)}} \\
		\hline
		& {\bf Oxford} & {\bf U.S.} & {\bf R.A.} & {\bf B.D.} & {\bf Oxford} & {\bf U.S.} & {\bf R.A.} & {\bf B.D.} \\
		\hline
{\bf RIConv} &       88.2 &       90.3 &       87.3 &       82.6 &       73.3 &       75.5 &       74.8 &       75.2 \\

{\bf SPRIN} &       89.5 &       92.1 &       88.2 &       83.7 &       79.2 &       78.5 &       78.7 &         80.0 \\

{\bf Ours} & {\bf 92.2} & {\bf 94.5} & {\bf 91.3} & {\bf 86.4} &   {\bf 81.0} & {\bf 83.2} & {\bf 83.3} & {\bf 80.4} \\

		\hline
		\hline
		{\bf } & \multicolumn{ 4}{|c}{{\bf Avg recall at top 1\%(w/ rotation)}} & \multicolumn{ 4}{|c}{{\bf Avg recall at top 1 (w/ rotation)}} \\
		
		\hline
		& {\bf Oxford} & {\bf U.S.} & {\bf R.A.} & {\bf B.D.} & {\bf Oxford} & {\bf U.S.} & {\bf R.A.} & {\bf B.D.} \\
		\hline
		 {\bf RIConv} &       88.3 &       90.2 &       86.8 &       83.2 &       73.4 &       76.1 &       74.9 &       76.2 \\
{\bf SPRIN} &       89.2 &       91.8 &       89.2 &       83.6 &         79.0 &       78.8 &         79.0 &       76.2 \\
{\bf Ours} & {\bf 92.2} & {\bf 93.8} & {\bf 91.4} & {\bf 86.3} & {\bf 81.1} & {\bf 83.3} &   {\bf 82.0} &   {\bf 80.0} \\

		\hline
	\end{tabular} } 
\caption{Quantitative comparison of our method with sate-of-the-art rotation-invariant baseline models at the baseline training setting. Top rows are results on the original datasets. Bottom rows are results when both submaps and query point clouds are randomly rotated along three axis.}
\vspace{-0.4in}
\label{rot_inv_baselines}
\end{table}

To see if our method can deal with more challenging cases, we randomly rotate the point clouds along the three axis following the evaluation setting of \cite{you2020pointwise,you2021prin}. Results on the baseline training setting are shown in Table \ref{tablebaseline}. The models' performances under non-rotated situation are also included.  We can find that when point clouds are not rotated (the original datasets), our method  achieves competitive results with existing methods, which means except to keep the rotation-invariant property, our model also retains the  ability of representing as much scene geometries and semantics in the global descriptor as possible. Then, when all point clouds are randomly rotated, our method achieves the state-of-the-art performance. Compared to the non-rotated situation, there is almost no performance drop. However, all other methods almost don't work at all.   The above analysis again demonstrates that our model is superior to other competitors.  Note again that rotation problems would frequently  happen in practical scenarios as shown in Fig. \ref{task}, therefore, according the above experimental results, our model has inherent advantages in ensuring recognition reliability.

\noindent
\textbf{Quantitative comparison with rotation-invariant baseline models:} Since existing place recognition models all fail to handle rotation problems. To further demonstrate the superiority of RPR-Net, we adopt RIConv \cite{zhang2019rotation} and SPRIN  \cite{you2021prin}, two state-of-the-art rotation-invariant backbones for classification and segmentation, as  $\varphi$ to learn rotation-invariant global descriptors.  Both baseline models are carefully tuned to achieve their best performance. Results on the baseline training setting are shown in Table \ref{rot_inv_baselines}. Though all models achieve rotation-invariant, our model significantly  outperforms RIConv and SPRIN.  We believe the superiority of our method comes from three aspects:  1) We propose three kinds of RIFs to inherit the rotation-invariant property as well as  extracting local geometry. 2) We propose an attentive module to learn powerful rotation-invariant convolutional kernels. 3) We propose a dense network architecture to better capture scene semantics. For results on the refined  training setting, please see the \textbf{Supplementary Materials}.

\begin{figure}
	\centering  
	\vspace{-0.35in}
	\includegraphics[width=0.7\textwidth]{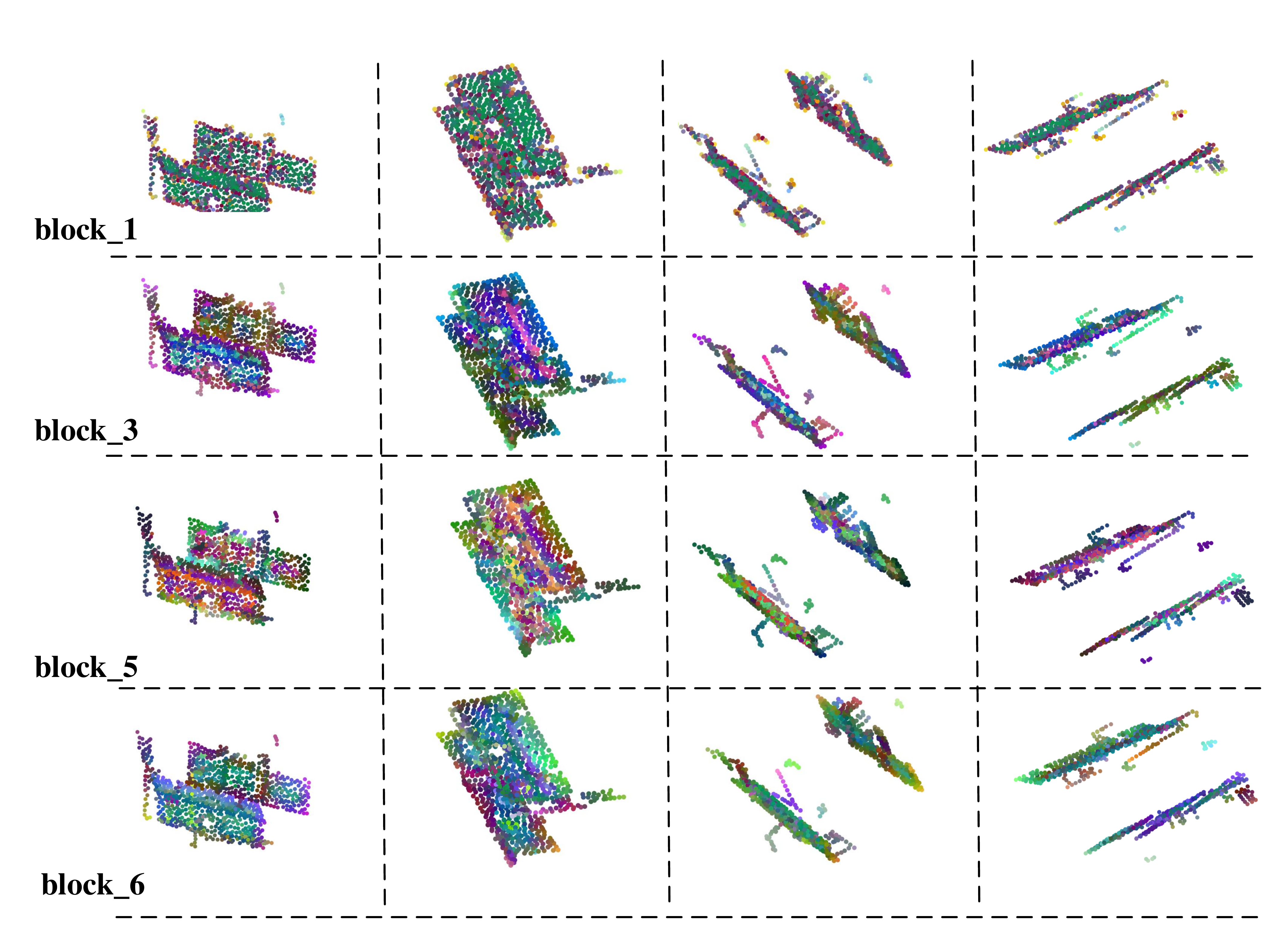}
	\vspace{-0.25in}
	\caption{T-SNE  visualization  of learned features.} 
	\label{visulfeatures}
	\vspace{-0.4in}
\end{figure}

\noindent
\textbf{Qualitative verification:} To verify if our model indeed learns point-wise rotation-invariant features, we visualize  learned features of different layers using  T-SNE in Fig. \ref{visulfeatures}. It can be seen that no matter how we rotate the same point cloud, the output point-wise features keep unchanged. It indicates that RPR-Net indeed learns strictly rotation-invariant features. We attribute this to the strong power of the proposed RIFs and the learned rotation-invariant convolutional kernels. 

\noindent
\textbf{Complexity analysis:} The total number of trainable parameters of our model is 1.1M, which is comparable to the current most light-weight models Minkloc3D (1.1M) and SVT-Net (0.9M). However, our model is much more robust towards rotation than them as discussed before. For running time, our model takes 0.283s to process a point cloud scan on a V100 GPU, which is relatively slow. Therefore, there is still much room for running time reduction. We leave solving the limitation as our future work.

\begin{table}
	\centering
	\resizebox{0.95\textwidth}{!}{
\begin{tabular}{c|c|c|c|c|c|c|c|c}
\hline
           & \multicolumn{ 4}{c|}{{\bf Avg recall at top 1\%  (w/o rot)}} & \multicolumn{ 4}{|c}{{\bf Avg recall at top 1\%  (w rot)}} \\
           \hline
           & {\bf Oxford} & {\bf U.S.} & {\bf R.A.} & {\bf B.D.} & {\bf Oxford} & {\bf U.S.} & {\bf R.A.} & {\bf B.D.} \\
\hline
{\bf w/o dense connection} &       90.1 &       91.4 &       90.4 &         84.0 &       90.3 &       91.6 &       89.8 &       83.4 \\

{\bf w/  NetVLAD} &       90.4 &       90.8 &       87.6 &       81.3 &       90.6 &       90.1 &       86.2 &       81.4 \\
{\bf Ours} & {\bf 92.2} & {\bf 94.5} & {\bf 91.3} & {\bf 86.4} & {\bf 92.2} & {\bf 93.8} & {\bf 91.4} & {\bf 86.3} \\
\hline
\end{tabular}  }

	\caption{Results about impact of network architecture.}
	\label{ablationnetwork}
	\vspace{-0.5in}
\end{table}

\subsection{Ablation Study}
In this subsection, we experimentally analyse the the necessity of each module and the correctness of the network designs. All experiments are conducted at the baseline training setting. The evaluation metric is average recall at top 1\%. Both results on randomly rotated along three axes and results on non-rotated situation are shown. The impact of the network architecture and the impact of ARIConv designs are investigated in this part. For more experiments, please see the \textbf{Supplementary Materials}.

\noindent
\textbf{Impact of network architecture:} The dense network architecture is one of the key componet of improving performance.  It plays an important role in learning high quality high-level semantic features. As shown in Table \ref{ablationnetwork}, when it is removed, the performance decreases a lot. Then, expect using GeM Pooling for aggregating global descriptors, many works \cite{uy2018pointnetvlad,zhang2019pcan,liu2019lpd} utilize NetVLAD \cite{arandjelovic2016netvlad} to achieve this function. Therefore, we also replace GeM Pooling with NetVLAD to compare their performances. Obviously, from Table \ref{ablationnetwork}, we find that though both methods can achieve rotation-invariant, GeM Pooling is a better choice because it demonstrates better overall performance.

\begin{table}
	\centering
	\vspace{-0.2in}
\resizebox{0.95\textwidth}{!}{
\begin{tabular}{c|c|c|c|c|c|c|c|c}

\hline
    {\bf } & \multicolumn{ 4}{c|}{{\bf Avg recall at top 1\%  (w/o rot)}} & \multicolumn{ 4}{|c}{{\bf Avg recall at top 1\%  (w/ rot)}} \\
\hline
    {\bf } & {\bf Oxford} & {\bf U.S.} & {\bf R.A.} & {\bf B.D.} & {\bf Oxford} & {\bf U.S.} & {\bf R.A.} & {\bf B.D.} \\
\hline
{\bf w/o GLRIF} & {\bf 92.7} &         92 &       90.4 &       85.5 & {\bf 92.7} &       91.7 &       91.1 &       85.6 \\

{\bf w/o ILRIF} &       46.9 &       59.1 &       59.8 &       48.7 &       46.6 &       59.5 &       58.5 &       49.2 \\

{\bf w/o  SS} &       91.4 &       94.1 &       89.7 &       85.1 &       91.4 &       93.6 &         90.0 &       85.3 \\

{\bf w/o attentive} &       92.3 &       92.7 &       90.6 &       85.4 &       92.2 &       93.1 &       89.9 &       85.6 \\

{\bf Ours} &       92.2 & {\bf 94.5} & {\bf 91.3} & {\bf 86.4} &       92.2 & {\bf 93.8} & {\bf 91.4} & {\bf 86.3} \\
\hline
\end{tabular}  
  }
 
	\caption{Results about impact of ARIConv.}
	\vspace{-0.4in}
	\label{ablationconvs}
	
\end{table}

\noindent
\textbf{Impact of ARIConv:} ARIConv is the most basic and pivotal component of our work. It core  units are the three kinds of RIFs and the attentive operation. In the Table \ref{ablationconvs}, we investigate  the impacts of them. First, we conclude that all three kinds of RIFs can hold the rotation-invariant property of the learned features, because no matter using what kinds of  RIFs, the place recognition performance doesn't decrease if we rotate the input point clouds. Second, we also find that the ILRIF is the strongest component since the performance greatly drops when it is removed. Meanwhile, the other two kinds of modules also contribute, though the contribution is not equal. Third, we note that the attentive module in ARIConv is also critical. When it is removed, the model's performances on the three indoor datasets decrease a lot, which means the generalization ability of the model is reduced. We also tried to directly mapping RIFs into high-dimensional features. However, in this case, the model fails to converge, therefore, we believe learning convolutional kernels is the right way of inheriting the rotation-invariant property of RIFs. In a word, all designs in ARIConv is necessary and effective.

\section{Conclusion}

In this paper, we propose a novel rotation-aware large scale place recognition model named RPR-Net, which focus on learning rotation-invariant global point cloud descriptors to solve the catastrophic collapse caused by rotation. We propose a novel ARIConv, which is equipped with three kinds of RIFs and an attentive module to conduct rotation-invariant convolution operation.  Taking ARIConv as a basic unit, a dense network architecture is constructed. Experimental comparison with state-of-the-art place recognition models and rotation-invariant models has demonstrated the superiority of our method. Our research may be potentially  applied to SLAM, robot navigation, autonomous driving, etc, to increase the robustness and the reliability of place recognition.

\section{Acknowledgements}
This work was supported by the National Social Science Major Program under Grant No. 20\&ZD161 and National Natural Science Foundation of China (NSFC) under Grant No. 62172421.

\bibliographystyle{splncs04}
\bibliography{egbib}

\begin{thebibliography}{10}
\providecommand{\url}[1]{\texttt{#1}}
\providecommand{\urlprefix}{URL }
\providecommand{\doi}[1]{https://doi.org/#1}

\bibitem{arandjelovic2016netvlad}
Arandjelovic, R., Gronat, P., Torii, A., Pajdla, T., Sivic, J.: Netvlad: Cnn
  architecture for weakly supervised place recognition. In: Proceedings of the
  IEEE conference on computer vision and pattern recognition. pp. 5297--5307
  (2016)

\bibitem{campos2021orb}
Campos, C., Elvira, R., Rodr{\'\i}guez, J.J.G., Montiel, J.M., Tard{\'o}s,
  J.D.: Orb-slam3: An accurate open-source library for visual,
  visual--inertial, and multimap slam. IEEE Transactions on Robotics  (2021)

\bibitem{chen2019clusternet}
Chen, C., Li, G., Xu, R., Chen, T., Wang, M., Lin, L.: Clusternet: Deep
  hierarchical cluster network with rigorously rotation-invariant
  representation for point cloud analysis. In: CVPR. pp. 4994--5002 (2019)

\bibitem{cohen2018spherical}
Cohen, T.S., Geiger, M., K{\"o}hler, J., Welling, M.: Spherical cnns. arXiv
  preprint arXiv:1801.10130  (2018)

\bibitem{dym2020universality}
Dym, N., Maron, H.: On the universality of rotation equivariant point cloud
  networks. arXiv preprint arXiv:2010.02449  (2020)

\bibitem{esteves2018learning}
Esteves, C., Allen-Blanchette, C., Makadia, A., Daniilidis, K.: Learning so (3)
  equivariant representations with spherical cnns. In: ECCV. pp. 52--68 (2018)

\bibitem{fan2020srnet}
Fan, Z., Liu, H., He, J., Sun, Q., Du, X.: Srnet: A 3d scene recognition
  network using static graph and dense semantic fusion. In: CGF. vol.~39, pp.
  301--311. Wiley Online Library (2020)

\bibitem{fan2021svt}
Fan, Z., Song, Z., Liu, H., He, J., Du, X.: Svt-net: A super light-weight
  network for large scale place recognition using sparse voxel transformers.
  arXiv preprint arXiv:2105.00149  (2021)

\bibitem{huang2017densely}
Huang, G., Liu, Z., Van Der~Maaten, L., Weinberger, K.Q.: Densely connected
  convolutional networks. In: CVPR. pp. 4700--4708 (2017)

\bibitem{kim2020rotation}
Kim, S., Park, J., Han, B.: Rotation-invariant local-to-global representation
  learning for 3d point cloud. arXiv preprint arXiv:2010.03318  (2020)

\bibitem{komorowski2021minkloc3d}
Komorowski, J.: Minkloc3d: Point cloud based large-scale place recognition. In:
  WACV. pp. 1790--1799 (2021)

\bibitem{li2020rotation}
Li, X., Li, R., Chen, G., Fu, C.W., Cohen-Or, D., Heng, P.A.: A
  rotation-invariant framework for deep point cloud analysis. arXiv preprint
  arXiv:2003.07238  (2020)

\bibitem{liu2019variance}
Liu, L., Jiang, H., He, P., Chen, W., Liu, X., Gao, J., Han, J.: On the
  variance of the adaptive learning rate and beyond. arXiv preprint
  arXiv:1908.03265  (2019)

\bibitem{liu2019densepoint}
Liu, Y., Fan, B., Meng, G., Lu, J., Xiang, S., Pan, C.: Densepoint: Learning
  densely contextual representation for efficient point cloud processing. In:
  CVPR. pp. 5239--5248 (2019)

\bibitem{liu2019lpd}
Liu, Z., Zhou, S., Suo, C., Yin, P., Chen, W., Wang, H., Li, H., Liu, Y.H.:
  Lpd-net: 3d point cloud learning for large-scale place recognition and
  environment analysis. In: CVPR. pp. 2831--2840 (2019)

\bibitem{maddern20171}
Maddern, W., Pascoe, G., Linegar, C., Newman, P.: 1 year, 1000 km: The oxford
  robotcar dataset. The International Journal of Robotics Research
  \textbf{36}(1),  3--15 (2017)

\bibitem{memon2020loop}
Memon, A.R., Wang, H., Hussain, A.: Loop closure detection using supervised and
  unsupervised deep neural networks for monocular slam systems. Robotics and
  Autonomous Systems  \textbf{126},  103470 (2020)

\bibitem{paszke2017automatic}
Paszke, A., Gross, S., Chintala, S., Chanan, G., Yang, E., DeVito, Z., Lin, Z.,
  Desmaison, A., Antiga, L., Lerer, A.: Automatic differentiation in pytorch
  (2017)

\bibitem{qi2017pointnet}
Qi, C.R., Su, H., Mo, K., Guibas, L.J.: Pointnet: Deep learning on point sets
  for 3d classification and segmentation. In: CVPR. pp. 652--660 (2017)

\bibitem{qi2017pointnet2}
Qi, C.R., Yi, L., Su, H., Guibas, L.J.: Pointnet++: Deep hierarchical feature
  learning on point sets in a metric space. Advances in neural information
  processing systems  \textbf{30} (2017)

\bibitem{radenovic2018fine}
Radenovi{\'c}, F., Tolias, G., Chum, O.: Fine-tuning cnn image retrieval with
  no human annotation. TPAMI  \textbf{41}(7),  1655--1668 (2018)

\bibitem{rao2019spherical}
Rao, Y., Lu, J., Zhou, J.: Spherical fractal convolutional neural networks for
  point cloud recognition. In: CVPR. pp. 452--460 (2019)

\bibitem{sun2020dagc}
Sun, Q., Liu, H., He, J., Fan, Z., Du, X.: Dagc: Employing dual attention and
  graph convolution for point cloud based place recognition. In: ICMR. pp.
  224--232 (2020)

\bibitem{sun2019srinet}
Sun, X., Lian, Z., Xiao, J.: Srinet: Learning strictly rotation-invariant
  representations for point cloud classification and segmentation. In: ACM MM.
  pp. 980--988 (2019)

\bibitem{uy2018pointnetvlad}
Uy, M.A., Lee, G.H.: Pointnetvlad: Deep point cloud based retrieval for
  large-scale place recognition. In: CVPR. pp. 4470--4479 (2018)

\bibitem{wang2019dynamic}
Wang, Y., Sun, Y., Liu, Z., Sarma, S.E., Bronstein, M.M., Solomon, J.M.:
  Dynamic graph cnn for learning on point clouds. ToG  \textbf{38}(5),  1--12
  (2019)

\bibitem{woo2018cbam}
Woo, S., Park, J., Lee, J.Y., Kweon, I.S.: Cbam: Convolutional block attention
  module. In: ECCV. pp. 3--19 (2018)

\bibitem{xia2021soe}
Xia, Y., Xu, Y., Li, S., Wang, R., Du, J., Cremers, D., Stilla, U.: Soe-net: A
  self-attention and orientation encoding network for point cloud based place
  recognition. In: CVPR. pp. 11348--11357 (2021)

\bibitem{you2020pointwise}
You, Y., Lou, Y., Liu, Q., Tai, Y.W., Ma, L., Lu, C., Wang, W.: Pointwise
  rotation-invariant network with adaptive sampling and 3d spherical voxel
  convolution. In: AAAI. vol.~34, pp. 12717--12724 (2020)

\bibitem{you2021prin}
You, Y., Lou, Y., Shi, R., Liu, Q., Tai, Y.W., Ma, L., Wang, W., Lu, C.:
  Prin/sprin: On extracting point-wise rotation invariant features. arXiv
  preprint arXiv:2102.12093  (2021)

\bibitem{yu2019spatial}
Yu, J., Zhu, C., Zhang, J., Huang, Q., Tao, D.: Spatial pyramid-enhanced
  netvlad with weighted triplet loss for place recognition. TNNLS
  \textbf{31}(2),  661--674 (2019)

\bibitem{yu2020deep}
Yu, R., Wei, X., Tombari, F., Sun, J.: Deep positional and relational feature
  learning for rotation-invariant point cloud analysis. In: ECCV. pp. 217--233.
  Springer (2020)

\bibitem{zhang2019pcan}
Zhang, W., Xiao, C.: Pcan: 3d attention map learning using contextual
  information for point cloud based retrieval. In: CVPR. pp. 12436--12445
  (2019)

\bibitem{zhang2019rotation}
Zhang, Z., Hua, B.S., Rosen, D.W., Yeung, S.K.: Rotation invariant convolutions
  for 3d point clouds deep learning. In: 3DV. pp. 204--213. IEEE (2019)

\end{thebibliography}


\begin{thebibliography}{1}
\providecommand{\url}[1]{\texttt{#1}}
\providecommand{\urlprefix}{URL }
\providecommand{\doi}[1]{https://doi.org/#1}

\bibitem{fan2021svt}
Fan, Z., Song, Z., Liu, H., He, J., Du, X.: Svt-net: A super light-weight
  network for large scale place recognition using sparse voxel transformers.
  arXiv preprint arXiv:2105.00149  (2021)

\bibitem{komorowski2021minkloc3d}
Komorowski, J.: Minkloc3d: Point cloud based large-scale place recognition. In:
  WACV. pp. 1790--1799 (2021)

\bibitem{liu2019variance}
Liu, L., Jiang, H., He, P., Chen, W., Liu, X., Gao, J., Han, J.: On the
  variance of the adaptive learning rate and beyond. arXiv preprint
  arXiv:1908.03265  (2019)

\bibitem{paszke2017automatic}
Paszke, A., Gross, S., Chintala, S., Chanan, G., Yang, E., DeVito, Z., Lin, Z.,
  Desmaison, A., Antiga, L., Lerer, A.: Automatic differentiation in pytorch
  (2017)

\end{thebibliography}
\newpage
\section{Supplementary Materials}
\subsection{More Implementation Details}
Here we introduce some additional implementation details about network training and testing to help readers understand our settings better and enhance the reproducibility of our method.

\noindent \textbf{Training settings} The raw point cloud contains 4096 points and we down-sample it into 1024 points to generate the seed point cloud. We use a dynamic batch sizing strategy like in \cite{komorowski2021minkloc3d} to generate training batches. We implement our model using  PyTorch \cite{paszke2017automatic} platform and optimize it using  the Radam optimizer \cite{liu2019variance}. Random jitter, random translation, random points removal and random erasing augmentation are adapted for data augmentation during training.  All experiments are conducted on a Tesla V100 GPU. 

\noindent \textbf{Dynamic batch sizing}: To make the training more stable, we use the dynamic batch sizing strategy proposed in MinkLoc3D during training. Specifically,  the training starts with a small batch size. Then, at the end of each epoch, we examine the average number of active triplets or  triplets producing non-zero loss. If the ratio of active triplets to all triplets in a batch falls below the predefined threshold, the batch size should be  increased by a expansion rate $r$.  Specifically, we count the number of active triplets, when it falls below 70\% of the current batch size, the batch is increased by 40\% until the maximum size of 96 elements is reached. The initial batch size is 16.

\noindent  \textbf{Define of positive/negative samples}: For a fair comparison, we use the same setting as all previous methods to define positive samples and negative samples. In training, point clouds are regarded as correct matches if they are at maximum 10m apart and wrong matches if they are at least 50m apart. In testing, the retrieved point cloud is regarded as a correct match if the distance is within 25m between the retrieved point cloud and the query scan.

\subsection{Loss Function}
Since the triplet loss has achieved superior performance in Minlok3D \cite{komorowski2021minkloc3d} and SVT-Net \cite{fan2021svt}, we follow them and use the same loss function to train our model for a fair comparison. The  triplet loss can be defined as:
\begin{equation}
L(f,f^{pos},f^{neg})=max\{d(f,f^{pos})-d(f,f_i^{neg})+m,0\}
\end{equation}
where $f$ is the descriptor of the query scan, $f^{pos}$ and $f^{neg}$ are the global descriptors of the corresponding positive sample and negative sample.  $m$ is a margin. $d()$ means the Euclidean distance between the two global descriptors in the latent feature space. The triplet loss plays a role of pushing the negative sample away from query scan as well as closing the distance between the query scan and the positive sample. Therefore, after the model is trained, descriptors extracted by it would be distributed in the feature space that best fit their position information in the 3D space and best fit their semantic contextual information. We also follow Minkloc3D and SVT-Net to use the batch-hard negative mining to build triplets during training.

\begin{table*}
	\centering
	\resizebox{0.95\textwidth}{!}{
	\begin{tabular}{c|c|c|c|c|c|c|c|c}
		\hline
		& \multicolumn{ 4}{|c|}{{\bf Avg recall at top 1\%  (w/o rotation)}} & \multicolumn{ 4}{|c}{{\bf Avg recall at top 1 (w/o rotation)}} \\
		\hline
		& {\bf Oxford} & {\bf U.S.} & {\bf R.A.} & {\bf B.D.} & {\bf Oxford} & {\bf U.S.} & {\bf R.A.} & {\bf B.D.} \\
		\hline
		{\bf PointNetVLAD} &       80.1 &       90.1 &       93.1 &       86.5 &       63.3 &       86.1 &       82.7 &       80.1 \\
	
		{\bf PCAN} &       86.4 &       94.1 &       92.3 &         87.0 &       70.7 &       83.7 &       82.3 &       80.3 \\
	
		{\bf LPD-Net} &       98.2 &       98.2 &       94.4 &       91.6 &         93.0 &       90.5 &       97.4 &       85.9 \\
		
		{\bf Minkloc3D} & {\bf 98.5} &       99.7 &       99.3 &       96.7 & {\bf 94.8} & {\bf 97.2} & {\bf 96.7} &         94.0 \\
	
		{\bf SVT-Net} &       98.4 & {\bf 99.9} & {\bf 99.5} & {\bf 97.2} &       94.7 &         97.0 &       95.2 & {\bf 94.4} \\
		
{\bf Ours} & 93.6 &  97.9 & 97.4 & 92.4 & 83.5 &    90.0 &  91.7 & 87.7 \\
		\hline
		\hline
		{\bf } & \multicolumn{ 4}{|c|}{{\bf Avg recall at top 1\%  (w/ rotation)}} & \multicolumn{ 4}{|c}{{\bf Avg recall at top 1  (w/ rotation)}} \\
		\hline
		& {\bf Oxford} & {\bf U.S.} & {\bf R.A.} & {\bf B.D.} & {\bf Oxford} & {\bf U.S.} & {\bf R.A.} & {\bf B.D.} \\
		\hline
		{\bf PointNetVLAD} &      5.1      &       4.4     &     4.6       &    3.8        &    1.6        &      1.7      &      2.2      &     2.1       \\
		
				{\bf PCAN} &    5.2        &     5.2      &       6.4     &     4.2       &   1.7  &    1.4    &    3.4        &     2.4       \\
	
		{\bf LPD-Net} &   8.6   & 6.0 &  5.0   &    5.1  &  2.5    &   2.8  &   1.4   &     2.9   \\
		
		{\bf Minkloc3D} &    15.2        &   18.6     &     15.5    &     17.5   &  6.3    &     7.9       &    8.7   &       12.1     \\
	
		{\bf SVT-Net}  &   14.5  &     17.6       &  14.1      &    15.8        &  5.9    &  8.3   &    8.3      &      10.3      \\
		
{\bf Ours} & {\bf 93.5} & {\bf 98.1} & {\bf 97.1} & {\bf 92.2} & {\bf 83.1} & {\bf 90.3} &   {\bf 92.0} & {\bf 87.5} \\
		\hline
	\end{tabular} } 
	\caption{Quantitative comparison of our methods with sate-of-the-art place recognition methods at the refined training setting. Top rows are results on the original datasets. Bottom rows are results when both submaps and query point clouds are randomly rotated.}
	\label{tablerefined}

\end{table*}

\begin{table*}
\centering
\resizebox{0.95\textwidth}{!}{
\begin{tabular}{c|c|c|c|c|c|c|c|c}
		\hline
		& \multicolumn{4}{|c|}{{\bf Avg recall at top 1\%(w/o rotation)}} & \multicolumn{ 4}{|c}{{\bf Avg recall at top 1 (w/o rotation)}} \\
		\hline
		& {\bf Oxford} & {\bf U.S.} & {\bf R.A.} & {\bf B.D.} & {\bf Oxford} & {\bf U.S.} & {\bf R.A.} & {\bf B.D.} \\
		\hline
{\bf RIConv} &       90.9 &       96.7 &       94.8 &       92.1 &       77.4 &       88.5 &       85.5 &       85.9 \\

{\bf SPRIN} &       92.2 &       97.3 &       95.6 &       92.2 &         80.0 &       89.2 &       88.9 &       82.2 \\

{\bf Ours} & {\bf 93.6} & {\bf 97.9} & {\bf 97.4} & {\bf 92.4} & {\bf 83.5} &   {\bf 90.0} & {\bf 91.7} & {\bf 87.7} \\
		\hline
		\hline
		{\bf } & \multicolumn{ 4}{|c}{{\bf Avg recall at top 1\%(w/ rotation)}} & \multicolumn{ 4}{|c}{{\bf Avg recall at top 1 (w/ rotation)}} \\
		
		\hline
		& {\bf Oxford} & {\bf U.S.} & {\bf R.A.} & {\bf B.D.} & {\bf Oxford} & {\bf U.S.} & {\bf R.A.} & {\bf B.D.} \\
		\hline
{\bf RIConv} &       91.3 &       87.6 &       95.2 &       91.8 &       77.6 &       87.6 &       86.8 &       86.2 \\
{\bf SPRIN} &       92.3 &       97.6 &       96.2 &       91.8 &       80.1 &       90.4 &       89.3 &       86.7 \\
{\bf Ours} & {\bf 93.5} & {\bf 98.1} & {\bf 97.1} & {\bf 92.2} & {\bf 83.1} & {\bf 90.3} &   {\bf 92.0} & {\bf 87.5} \\

		\hline
	\end{tabular}  }
\caption{Quantitative comparison of our method with sate-of-the-art rotation-invariant baseline models at the refined training setting. Top rows are results on the original datasets. Bottom rows are results when both submaps and query point clouds are randomly rotated along three axis.}
\label{rot_inv_refined}
\end{table*}

\subsection{More quantitative comparison}
In Table \ref{tablerefined},  we compare our method with previous place recognition methods under the refined training setting. Performances on both rotated and non-rotated situations are shown. The rotation level is unlimited. We can find that when point clouds are not rotated (the original datasets), our method  achieves competitive results with existing methods, which means except to keep the rotation invariant property, our model also retains the feature's ability to represent the geometric and semantic context of the scene as much as possible. Then, when all point clouds are randomly rotated, our method achieves the state-of-the-art performance. Compared to non-rotated situation, there is almost no performance drop. However, all other methods almost don't work at all.   The above analysis again demonstrates that our model is superior to other competitors. 

In Table \ref{rot_inv_refined}, we also compare our method with other strong rotation-invariant baseline models at the refined training setting, it can be found that our method significantly outperforms other rotation-invariant models.

In a world, in the refined training setting, our method demonstrates the same conclusion as in the baseline training setting, which is that our method can learn discriminative strictly rotation-invariant global descriptors to promote the large scale place recognition performance. 

\subsection{More ablation study}
Except the experiments we exhibit in the paper, we conduct more ablation study about  hyper parameters selection in this section. 

\noindent  \textbf{Impact of the number of local neighbors:} The  number of local neighbors $K$ in each local group is very  important for  extracting RIFs because it determines the receptive field of the model. As results in Table \ref{ablationneighbors} indicate, $K$ can neither be too large nor too small. Setting $K$ as 32 is the best choice.  This is  reasonable. When $K$ is too small, the model cannot fully perceive the local information of the point cloud. When $K$ is too large, it is easy to cause over-fitting or introduce noise.

\noindent  \textbf{Impact of the number of the descriptor's dimension:} The final  factor we find that may  influence the performance of the model is the dimension of the final output point cloud descriptor. In Table \ref{ablationdimension}, we investigate its impact. In order to avoid sabotaging our dense network architecture, we choose to change the final dimension by adding a new ARIConv layer after ARIConv block\_6. Therefore, the results of row "l=256" is quite different with the result of our model in other tables. We can conclude from Table \ref{ablationdimension} that larger $l$ would bring performance gain. However, with $l$ increases, the model size increases dramatically. Therefore, we set $l$ as 256 for a better trade-off.

\begin{table}
	\centering
\begin{tabular}{c|c|c|c|c|c|c|c|c}
\hline
    {\bf } & \multicolumn{ 4}{c|}{{\bf Avg recall at top 1\%  (w/o rot)}} & \multicolumn{ 4}{|c}{{\bf Avg recall at top 1\%  (w/ rot)}} \\
\hline
    {\bf } & {\bf Oxford} & {\bf U.S.} & {\bf R.A.} & {\bf B.D.} & {\bf Oxford} & {\bf U.S.} & {\bf R.A.} & {\bf B.D.} \\
\hline
{\bf k=16} &       91.7 &         94.0 &       88.3 &       85.7 &       91.7 &       93.6 &       88.9 &         86.0 \\

{\bf k=24} & {\bf 92.8} &         93.0 &         90.0 &       85.5 & {\bf 92.8} &       92.6 &       89.6 &       85.8 \\

{\bf k=48} &       92.2 &         93.0 &       90.7 &       85.3 &       92.1 &       93.4 &       90.1 &       85.2 \\

{\bf k=64} &       90.8 &       92.7 &       91.3 &       85.1 &       90.7 &       92.7 &       91.1 &       85.3 \\

{\bf K=32} &       92.2 & {\bf 94.5} & {\bf 91.3} & {\bf 86.4} &       92.2 & {\bf 93.8} & {\bf 91.4} & {\bf 86.3} \\
\hline
\end{tabular}

	\caption{Results about impact of number of local neighbors, where "rot" means "rotation".}
	\label{ablationneighbors}
\end{table}

\begin{table}
	\centering
\begin{tabular}{c|c|c|c|c|c|c|c|c}
\hline
    {\bf } & \multicolumn{ 4}{c|}{{\bf Avg recall at top 1\%  (w/o rot)}} & \multicolumn{ 4}{|c}{{\bf Avg recall at top 1\%  (w/ rot)}} \\
\hline
    {\bf } & {\bf Oxford} & {\bf U.S.} & {\bf R.A.} & {\bf B.D.} & {\bf Oxford} & {\bf U.S.} & {\bf R.A.} & {\bf B.D.} \\
\hline
{\bf l=128} &       92.6 &       92.3 &       89.3 &       84.6 &       92.5 &       91.8 &       90.1 &       85.1 \\

{\bf l=256} &       92.3 &         93.0 &       91.3 &       85.4 &       92.4 &       93.1 &       91.4 &       84.9 \\

{\bf l=384} &       92.8 &       94.1 &       89.4 &       85.3 &       92.9 &       93.7 &         90.0 &       85.6 \\

{\bf l=512} &   {\bf 93.0} & {\bf 94.1} & {\bf 91.3} & {\bf 86.3} & {\bf 93.1} & {\bf 94.3} & {\bf 91.5} & {\bf 86.9} \\
\hline
\end{tabular}  

	\caption{Results about impact of number of descriptor dimension, where "rot" means "rotation".}
	\label{ablationdimension}
\end{table}

\subsection{Limitation and Future Work}
Though our RPR-Net has achieved superior performance when dealing with rotation problems, there are some limitations that should be tackled in the future.

 First,  the overall recognition accuracy  of our  RPR-Net at the non-rotated situation is not as good as other state-of-the-art methods like SVT-Net, though RPR-Net significantly outperforms these methods when solving rotation problems.  We need to investigate how to learn stronger rotation-invariant convolutions that can extract scene geometry and semantics better to solve the problem. That is to say that there is still much room for the overall performance improvement except for learning the rotation-invariant property. 

Second, though our model only contains minimal model parameters, the running time is long (0.283s). That is caused by the \emph{torch.einsum} operation in the network for learning convolutional kernels and conducting the convolution operation. Therefore, we have to design more computational efficient operations to reduce running time. We leave this to future work.
\end{document}